%% file: main.tex
\documentclass{article} 
\usepackage{iclr2025_conference,times}
\usepackage{booktabs}
\usepackage{algorithm}
\usepackage{algorithmicx}
\usepackage{graphicx}
\usepackage{sidecap}
\usepackage[noend]{algpseudocode}
\usepackage[breakable]{tcolorbox}

\input{math_commands.tex}

\usepackage{wrapfig}
\usepackage{hyperref}
\usepackage{url}

\title{ScreenWriter: Automatic Screenplay Generation and Movie Summarisation}

\author{Louis Mahon \thanks{ \url{https://lou1sm.github.io/louismahonville/}}\\
School of Informatics\\
University of Edinburgh\\
\texttt{lmahon@ed.ac.uk} \\
\And
Mirella Lapata \\
School of Informatics\\
University of Edinburgh\\
}

\usepackage[english]{babel}
\usepackage{ragged2e}

\iclrfinalcopy 
\begin{document}

\maketitle

\begin{abstract}

The  proliferation of creative video  content has driven demand for textual descriptions or summaries that allow users to recall key plot points or get an overview without watching. The volume of movie content and speed of turnover motivates automatic summarisation, which is nevertheless challenging, requiring identifying character intentions and very long-range temporal dependencies. The few existing methods attempting this task rely heavily on textual screenplays as input, greatly limiting their applicability. In this work, we 
propose the task of automatic screenplay generation, and a method, ScreenWriter, that operates only on video and produces output which includes dialogue, speaker names, scene breaks, and visual descriptions. ScreenWriter introduces a novel algorithm to segment the  video into scenes based on the sequence of visual vectors, and a novel method for the challenging problem of determining character names, based on a database of actors' faces. We further demonstrate how these automatic screenplays can be used to generate plot synopses with a hierarchical summarisation method based on scene breaks. 
We test the quality of the final summaries on the recent MovieSum dataset,  which we augment with videos, and show that they are superior to a number of comparison models which assume access to goldstandard screenplays. 
\end{abstract}

\section{Introduction}

Thanks to the proliferation of streaming services and digital content
providers, a large number of movies are being released and made
available every year. Automatic approaches to understanding and
summarising their content are paramount to enabling users to browse or
skim through them, and quickly recall key plot points, characters,
and events without the need to re-watch from the beginning. Aside from
practical utility, movie summarisation is an ideal testbed for
real-world video and natural language understanding. Movies are often
based on elaborate stories, with non-linear structure and multiple
characters whose emotions and actions are often not accompanied with
verbal cues. The summarisation task requires identifying beliefs and
intentions of characters as well as reasoning over
very long narratives (there are usually hundreds of pages in a
transcript and tens of thousands of frames in a video), involving multiple
modalities.


Most previous work on movie summarisation has focused primarily on the
textual modality, under the assumption that screenplays (or at least
human-authored transcripts) are readily available
\citep{gorinski-lapata-2015-movie,chen-etal-2022-summscreen,agarwal-etal-2022-creativesumm}. Movie summarisation
is commonly viewed as a type of long-form summarisation, a task which
has improved dramatically in recent years. This is partly due to large models
with longer context windows, and the design of methods specifically
for this task, such as dividing the input into chunks and forming a
hierarchical summarisation structure \citep{chen2023walking,
  pang-etal-2023-long} or summarising chunks iteratively
\citep{chang2023booookscore}. Video understanding through
generating descriptions for videos, has also received much attention,
largely independently from long-form summarisation, focussing instead
on short video clips lasting a few minutes
\citep{tapaswi2016movieqa,lei-etal-2018-tvqa,Rafiq:ea:2023}. Two
notable exceptions are \citet{papalampidi2023hierarchical3d} and
\citet{mahon-lapata-2024-modular} who consider textual \emph{and} visual
modalities, aiming to summarise movies and television shows,
respectively. However, they both require a textual transcript with
information about scene breaks, and character names prefacing the
lines actors speak. The assumption of having available written
transcripts and/or screenplays is unrealistic, as most video providers
do not have access to screenplays unless they have produced the
content themselves. Additionally, there are often differences between
original screenplays, and what is acted and shown in the movie.

In this paper, we address the challenging tasks of generating
automatic screenplays and plot summaries of movies by leveraging only
video and audio input, without relying on transcripts or text.
%
%
%
As a result, our method, ScreenWriter is applicable to the large
quantity of movie videos which exist without accompanying screenplays
or human-annotated transcripts. A naive solution is to apply a
speech-to-text model from which we could then build a rudimentary
screenplay automatically. Using speech-to-text off the shelf for movie
understanding is problematic for several reasons. Firstly, it does not
contain scene breaks, which can provide useful structural and semantic
information
\citep{gorinski-lapata-2015-movie,papalampidi-etal-2019-movie}. Secondly,
it does not contain descriptions of events outside of dialogue such as
characters fighting or kissing. And perhaps most importantly, it lacks
character names. Speaker diarization aims to determine whether
different utterances are spoken by the same speaker, but this speaker
is still indicated with an arbitrary ID, rather than a character name.

ScreenWriter attempts to reconstruct screenplays and their sctructure
as accurately as possible. It eschews the above mentioned pitfalls by
segmenting the video into scenes using a novel parameter-free method
based on the minimum description length principle coupled with a
dynamic programming search algorithm. It associates speaker IDs to
character names through a novel method which exploits a database of
actor face images, coupled with the name of the character they played
(which is readily available online). By computing distances between
face feature vectors in the same scene as the utterance, we formulate
the task as an instance of the linear sum assignment problem, which
can then be solved efficiently by known algorithms. Additionally, we
add event descriptions (as an approximation to action lines which
describe what the camera sees) using an image-to-text model on
keyframes detected in each scene. We further demonstrate that
automatically generated screenplays are good enough to produce textual
summaries: we condense each scene independently and then fuse the
information from all scenes into a final synopsis using
zero-shot prompting.

ScreenWriter, and the accompanying summarisation method, is therefore
modular, based on a number of interacting components that separately
solve different subtasks. This design differs from recent work
\cite{song2024moviechat} proposing  to  modify transformer memory
in order to handle longer video sequences. Scaling such end-to-end models to
full-length movies remains a significant challenge due to memory
constraints and the complexity of extracting useful information from
large inputs.\footnote{At 1,024 $\times$ 1,024 frame size, and 10 frames per
second, a 75min movie would consume over 500GB when represented as a
4d 32-bit float tensor.} 
Our approach, which breaks the task into subproblems which are then
solved separately, mitigates this issue and allows for more efficient
processing and summarisation. In summary, our contributions include:

\begin{itemize}
    \item Proposing the new task of screenplay generation from video
      (and audio), and a modular framework, ScreenWriter, to address
      this task;
    \item A novel, parameter free method for detecting scene boundaries in video;
    \item A novel algorithm for detecting character names based on a database of actors' faces;
    \item A-proof-of-concept summariser which takes   automatic screenplays from
      ScreenWriter as input and produces textual synopses following a
      hierarchical method with zero-shot prompting. 

\end{itemize}

\section{Related Work} \label{sec:related-work}

\paragraph{Video Understanding}
The problem of generating descriptions for videos has received
significant attention in the literature. Traditional video description
approaches often extract features from individual frames and fuse them
into a single feature vector to generate a textual description
\citep{zhang2021open, pan2020spatio, ye2022hierarchical}. SwinBERT
\citep{lin2022swinbert} introduces an end-to-end video network that
samples frames densely, avoiding the need for image-based
encoders. Similarly, \citet{lei-etal-2020-mart} use a memory-augmented
vision transformer to generate descriptions for short
videos. 
In the area of video segmentation, \citet{souvcek2020transnet}
introduce TransNet~v2, a deep learning model for shot transition
detection. However, shot boundaries are easier to detect than scene
boundaries because there is a more striking pixel-level discontinuity.

Despite this progress, many popular video
description datasets \citep{chen2011collecting, xu2016msr} contain
videos that are only ${\sim}10$s in length, which are much shorter
than full-length movies. YouCook \citep{ZhXuCoCVPR18}, a dataset with
longer videos (about 5 minutes on average), remains insufficient for
capturing the complexities of movie understanding and
summarisation. Egoschema \citep{mangalam2023egoschema} explores video
understanding for clips of $\sim$3 minutes, focusing on human activity
and behavior, but this too is far shorter and more constrained than
full movie summarisation.
Some methods have started to tackle movie content, but in the form of
much shorter clips. \citet{wu2021towards} use YouTube movie clips for
small tasks like classifying movie year or genre, but  lack a
comprehensive approach for understanding/summarising an entire movie.
\citet{chen2023movies2scenes} propose Movies2Scenes, a method that
uses movie metadata to learn scene representations, though it relies
on predefined scenes based on shot transitions rather than
semantically meaningful boundaries. \citet{han2024autoad} address
the problem of assigning names to characters within a local movie
clip, using a "character bank" of images which is then used to prompt
a LLM for character names. This differs from our method in the use of a
LLM for name assignment, and in that we are assigning names
to  speaker IDs that arise from speaker diarization, rather than to
dialogue directly. Moreover, our assignment concerns the entire movie,
rather than a short snippet.

\paragraph{Long-form summarisation}
Recent work has started to address the task of summarising much longer
videos. \cite{papalampidi2021movie} describe a method for summarising
full-length movies by creating shorter videos containing their most
informative scenes which they assume to be `turning points' (i.e.,~key
events in a movie).  \citet{papalampidi2023hierarchical3d} produce
text summaries of TV show episodes, by converting visual features into
embeddings alongside word embeddings from the
transcript. \citet{mahon-lapata-2024-modular} also summarise TV show episodes
from video and text input, by converting the video to text, and then
treating it as a text-only problem.  Our method differs from all of
these approaches in that it only requires video as input, tackling the
more challenging task of understanding who is speaking and what they
are doing (this  information is taken for granted in screenplays and
human-authored transcripts).


\paragraph{Movie Datasets} Datasets containing videos for full-length feature
movies are few and far between possibly due to copyright restrictions
and the computational overhead of storing and processing lengthy
videos. \cite{papalampidi:ea:2024} introduce TRIPOD, a dataset
consisting of screenplays, their movies, and corresponding trailers,
however no summaries or synopses are provided. MovieNet
\citep{huang2020movienet} contains summaries and other annotations for
movies, but not the videos themselves, which are specifically the
focus of the present work.\footnote{The MovieNet website claims that
videos will be released soon, but at the time of writing, despite
several years wait, they have not appeared.}
\cite{papalampidi2023hierarchical3d} release SummScreen$^{3D}$ a
dataset consisting of soap opera episodes, their (crowd-sourced)
transcripts, and summaries. However, these episodes are not
self-contained, as they typically refer to events or characters from
previous episodes. In addition, they tend to be relatively short, and dialogue-heavy, with recurring scene locations. 
In this work, we wish to explore longer videos of 70+ minutes, and
move away from the niche domain of soap operas. In experiments, we use
the recent MovieSum dataset \citep{saxena2024moviesum} which contains
screenplays and their Wikipedia plot summaries.  We augment this
dataset with videos that we privately procure for the test set.

\section{ScreenWriter} \label{sec:screenwriter}
\begin{SCfigure}[1][ht]
    \centering
    \includegraphics[width=0.51\textwidth]{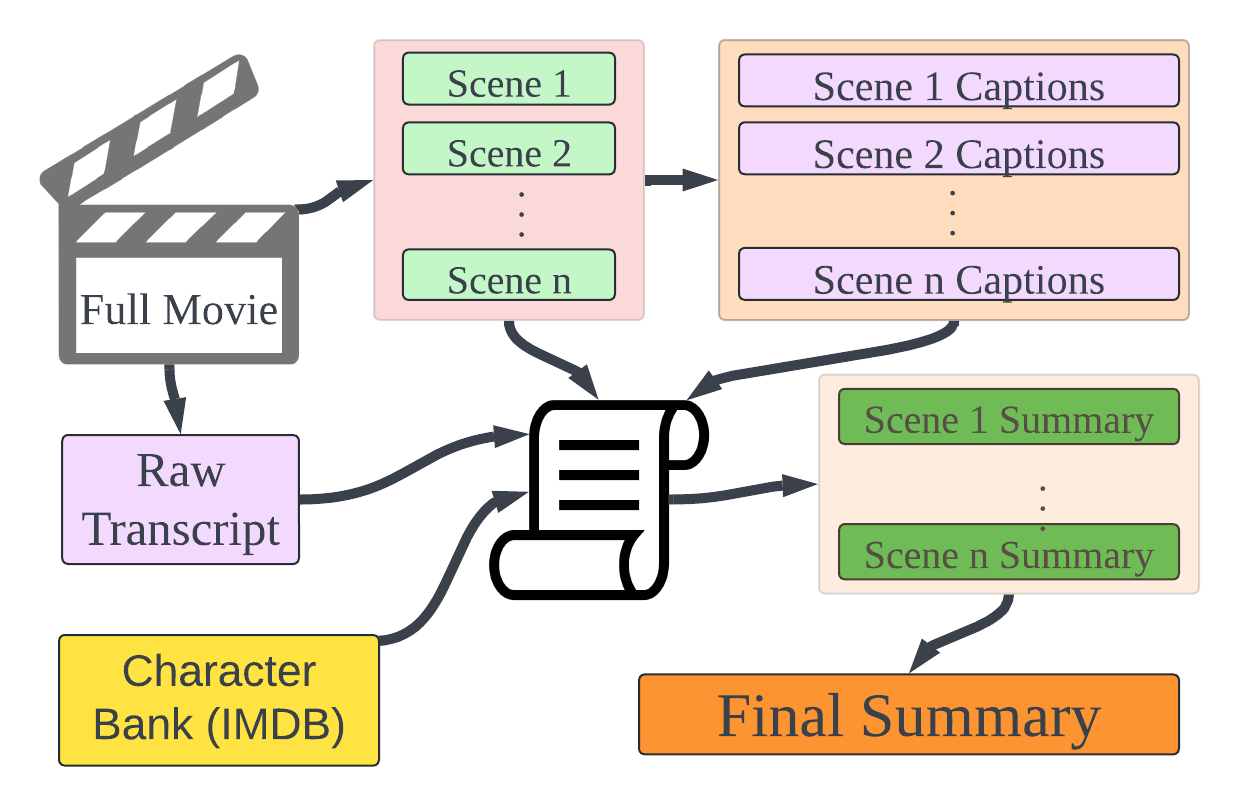}
    \caption{Screenwriter produces a screenplay from
      the input video/audio (top left), by first extracting the raw transcript,
      segmenting the video into scenes, and generating visual descriptions
      from each scene (top right). Then from these outputs, and the character bank (bottom left), it assigns character names
      and generates the screenplay (centre). Given the generated
      screenplay, we can then produce summaries for each scene (centre right), and fuse 
      into a single summary for the entire movie
      (bottom right).}
    \label{fig:method}
\end{SCfigure}

Figure~\ref{fig:method} provides a graphic depiction of
ScreenWriter. As mentioned earlier, the input to ScreenWriter is the
video (including audio) for a full movie. We extract keyframes
from the movie and use our novel scene segmentation algorithm (see
Section~\ref{subsec:scene-detection}) to partition the resulting sequence of frames into different
scenes. In parallel, a text-to-speech model with speaker diarization
yields a transcript with numeric speaker IDs instead of character
names. For each movie, we assume access to a database consisting of
actors faces and their character names (such a database can be easily
constructed by scraping the faces of the characters from the movie's
IMDB page).  We extract facial features from each keyframe in each
scene, and from each scraped character image, and use our novel name
assignment algorithm to replace (some of) the numeric IDs in the
transcript with actual character names (see Section~\ref{alg:character-name-assignment}). Finally, to insert visual
information into the screenplay about what is happening on camera, we
extract textual descriptions from three evenly spaced keyframes in
each scene using a multimodal large language model.
We now describe  these components in detail.


\subsection{Transcript Generation}

The first step in ScreenWriter is to extract the audio from the movie,
using an automatic transcription model with speaker diarization \citep{bain23_interspeech}. This
produces a transcript with each utterance marked by a speaker ID. The speaker ID is the same for the same character
throughout the movie (up to the accuracy of the diarization), but the
character names are lacking. We also extract key frames from the
entire video, and store the timestamps of these keyframes. These will
be used to compute speaker names and visual information, which we will
add to the diarized transcript to form a screenplay. It is also on
this sequence of keyframes that the scene breaks are computed. There
is an average of around 1,000 detected keyframes for a feature-length
movie.

\subsection{Scene Detection} \label{subsec:scene-detection}
ScreenWriter extracts visual features from each keyframe, and applies
a novel algorithm to segment the resulting sequence of visual feature
vectors. There are two parts to our algorithm: the \emph{definition of a
cost} for a particular partition into scenes, and the \emph{search} for the
partition that minimizes this cost. The first part, the cost
definition, is formulated using the minimum description length principle,
which claims the correct representation of the data is the one using
the fewest bits. We assume that the vectors for each scene are encoded
with respect to their collective mean. That is, for each scene in the
given partition, we calculate the mean of all vectors in that scene,
and hence, the probability of each vector, $p(v)$, under the
multivariate normal distribution with this mean. To reduce run time,
we use a single fixed covariance calculated from the entire sequence
of vectors. The Kraft-McMillan inequality \citep{kraft1949device,
  mcmillan1956two} then determines that under the optimal encoding,
the number of bits needed to represent $v$ is $-\log_2{p(v)}$. The sum
of this value across all $N$ vectors $v$ in the video, plus the number
of bits to represent the mean vectors themselves, gives the total
bitcost for a given partition. The mean vectors require $dm$~bits,
where $d$~is the dimensionality,
 and $m$ is the floating point precision, for which we use the
 standard of 32. Partitions with more scenes require more bits for the
 mean vectors, but also have mean vectors that better cover the
 keyframe features, leading to decreased $-\log_2{p(v)}$ on
 average. This trade-off encourages a partition with neither too few
 nor too many scene breaks.  

The second part, the search for the minimizer of the above cost, can be solved exactly using dynamic programming. Let $B(i,j)$ be the cost of having a single scene that runs from keyframes $i$ to $j$, and let $C(i,j)$ be the minimum cost of all keyframes from $i$ to $j$, under all possible partitions. Then we have the recurrence relation
\begin{equation} \label{eq:scene-cost-DP}
    C(i,j) = \min_{i \leq k \leq j} B(i,k) + C(k,j)\,.
\end{equation}
Thus, we can compute the optimal partition by iteratively computing and caching $C(i,N)$ for \mbox{$i=N-1, \dots, 0$}. This is guaranteed to find the global optimum. It runs in $O(N^2)$, but by imposing a fixed threshold of the maximum number $L$ of keyframes in a scene, this becomes $O(N)$. In our experiments, we find that setting $L=50$ produces no change in the solution. The full scene detection method is shown in Algorithm \ref{alg:scene-detection}.

\begin{algorithm}[t]
\caption{Scene Boundary Detection}
\label{alg:scene-detection}
\begin{footnotesize}

\begin{algorithmic}[1]
\State \textbf{Input:} Video file
\State Extract keyframes, $kf_0, \dots, kf_N$
\State Extract visual features $v_0, \dots, v_N$ from each keyframe
\State $L \gets 50$ \Comment{maximum scene length}
\State $B \gets $ $N \times N$ empty matrix \Comment{$B[i,j]$ will hold the cost of a scene from $v_i$ to $v_j$}
\State $\Sigma \gets $ empirical covariance matrix of $v_0, \dots, v_N$ 
\State $d \gets $ dimensionality of $v_i$
\State $m \gets $ floating point precision of $v_i$
\vspace{10pt}

\State \textbf{Cost Definition:} Compute and store costs for all possible scenes
\For{$0 \leq i \leq N-L$}
    \For{j=i, \dots, i+L}
        \State $\mu_{i,j} \gets \frac{1}{j-i}\sum_{k=i}^j v_k$
        \State $C \gets dm$
        \For {$k=i,\dots, j$}
            \State $p(v_k) \gets \frac{1}{(2\pi)^{d/2} |\Sigma|^{1/2}} \exp\left( -\frac{1}{2} (v_k - \mu_{i,j})^\top \Sigma^{-1} (v_k - \mu_{i,j}) \right)$
            \State $C \gets C - \log{p(v_k)}$
        \EndFor
        \State $B[i,j] \gets C$
    \EndFor
\EndFor
\vspace{10pt}

\State \textbf{Search:} Minimize the bitcost by dynamic programming
\State $C \gets B$ \Comment{will hold optimal costs}
\State $P \gets $ $N \times N$ matrix of empty sets \Comment{will hold optimal partitions}
\For{i=N-1, \dots, 0}
    \For{$j=i, \dots, \min(N, i+L)$}
        \If{$B[i,j] + C[j,N] < C[i,N]$}
            \State $C[i,N] \gets B[i,j] + C[j,N]$
            \State $P[i,N] \gets \{j\} \cup P[j,N]$
        \EndIf
    \EndFor
\EndFor
\vspace{10pt}

\State \textbf{Output:} Optimal scene partition, $P[0,N]$
\end{algorithmic}
\end{footnotesize}
\end{algorithm}

\subsection{Character Name Identification} \label{subsec:name-assignment}


To replace the arbitrary speaker IDs with character names, we first
create a database of images of actors' faces paired with the name of
the character they played from the IMDB movie page.
As some of these images may contain multiple faces, or
no faces, or even an entirely different character, we filter them to ensure a higher proportion contain only the face of
the correct character, keeping only images with exactly one detected face, and for which the detected gender matches the name gender. 
Finally, we verify the
faces in all pairs of remaining images against each other, using the
DeepFace\footnote{\url{https://github.com/serengil/deepface}} library, to create a graph where images are
connected if and only if they are verified as being the same person,
and then exclude all images that are not part of the largest
clique. In total, we filter out about 40\% of images on average. This
produces a name bank of character names paired with a set of images of
the face of that character.

\begin{figure}[t]
    \centering
    \includegraphics[width=\linewidth]{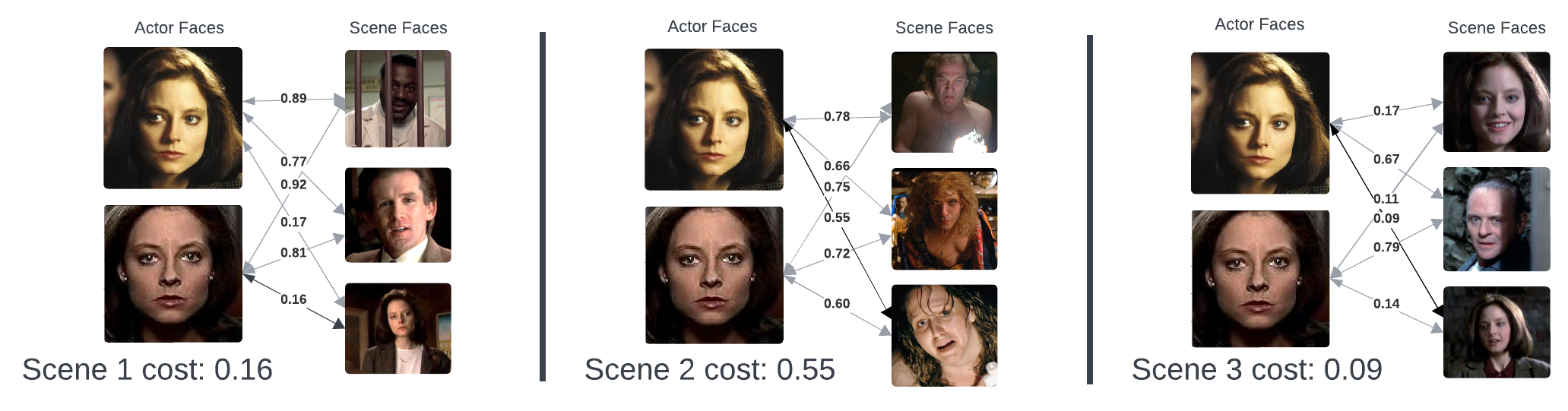}
    \caption{Computing the cost of assigning the character Clarice Starling (Jodie Foster) to three different scenes of \textit{The Silence of the Lambs} (1991). After computing the cost of assigning a character to a each scene, we then compute the cost of assigning a character to a speaker ID as the mean of the cost of assigning them to all scenes that speaker ID appears in.}
    \label{fig:scene-assignment-cost}
\end{figure}

For each scene, 
and for each character in our name bank, we define the cost of putting that character name in that scene as the minimum distance between an image of that character`s face, and a face detected in any keyframe from the scene. The distance is the Euclidean distance of the DeepFace feature vectors. This avoids the incorrect assumption that the character speaking must be in shot, and instead makes the much weaker assumption that a character speaking must appear directly at some point in the scene.
Thus, if we are considering assigning the character Clarice Starling to scene~3, then we compute the distance between the face feature vectors for all scraped images of the actor Jodie Foster in that role, and the face feature vectors of all faces detected in any keyframe in scene~3; the smallest distance is the cost of assigning Clarice Starling to scene 3. Computing the distance between vectors is extremely fast, taking $<$1s for all considered assignments on the entire movie, and the feature vectors can be cached after being extracted once. An example of this cost computation is shown in Figure \ref{fig:scene-assignment-cost}. 
Using this cost, we define the cost of assigning each character to each speaker ID, as the sum of assigning that character to all scenes that that speaker ID appears in
For example, if Speaker18 appears in scenes 1 and 3 but not 2, then the cost of assigning Clarice Starling to Speaker18 is the mean of the cost of assigning Clarice Starling to scenes 1 and 3. 
This allows us to treat the name-speaker ID assignment problem as an instance of the linear sum assignment problem, which can be solved efficiently using the Kuhn-Munkres algorithm \citep{kuhn1956variants,munkres1957algorithms}. 

Specifically, we define a matrix $S$ whose $i,j$th entry is the cost of assigning speaker $j$ to name $i$. Let $m$, $n$, and $k$ be the numbers of character names in the database, scenes in the movie, and unique speaker IDs in the transcript. Using matrix notation, we can then write $S = AB$, where $A$ is the $m \times n$ speaker ID-scene cost matrix, whose $i,j$th entry is the cost of assigning speaker $j$ to scene $i$, and $B$ is a $n \times k$ matrix whose $i,j$th entry is $1/a$ if speaker ID $j$ appears in scene $i$, where $a$ is the number of scenes speaker $j$ appears in, and 0 otherwise. 
Because speaker diarization is imperfect and often mistakenly splits the same character into multiple IDs, we duplicate each matrix column three times, which allows up the three different speaker IDS assigned to the same character name. We also define a cost of leaving a SpeakerID unassigned as expected value of the cost of assigning a random speaker ID to a random character, which means that an ID remains unassigned if it is no closer to any character than a random speaker ID and character are to each other.
The full name-assignment method is shown in Algorithm \ref{alg:character-name-assignment}.

\begin{algorithm}[t]
\caption{Character Name Assignment to Speaker IDs}
\label{alg:character-name-assignment}
\begin{footnotesize}
\begin{algorithmic}[1]
\State \textbf{Input:} Transcript with speaker IDs, keyframes split into $n$ scenes, IMDB
\vspace{10pt}

\State \textbf{Obtain actor face images:}
\State $\mathcal{A} \gets $ empty list
\For{each actor/character $A$ appearing on the IMDB page for the movie}
    \State scrape the set $A_f$ of all available images of $A$
    \State remove from $A_f$, all images without exactly one detected face, or with face-name gender mismatch
    \State form graph $G=(A_f, E)$, where $E = \{(a_1,a_2) \in A_f \times A_f | \text{isVerified}(a_1,a_2)\}$
    \State $A_f \gets $ largest clique in $G$
    \State append $A_f$ to $\mathcal{A}$
\EndFor
\For{each scene $j=1, \dots, n$}
    \State Form $D_j$, the set of all faces across all keyframes of the scene
\vspace{10pt}

\EndFor
\State \textbf{Assign character names to scenes:}
\State $C_1 \gets$ $n \times m$ empty matrix, where $m$ is the length of $\mathcal{A}$
\For{$i=1, \dots, m$}
    \State $A_f \gets \mathcal{A}[i]$
    \For{each scene $j=1, \dots, n$}
        \State $C_1[i,j] \gets min_{a \in A_f, b \in D_j} d(a,b)$ \Comment{$d(\cdot)$ from Deepface vectors}
    \EndFor
\EndFor
\vspace{10pt}

\State \textbf{Assign character names to speaker IDs:}
\State $C_2 \gets$ $k \times m$ empty matrix, where $k$ is the number of unique speaker IDs
\For{$i=1, \dots, m$}
    \For{each speaker ID $l=1, \dots k$}
        \State $C_2[i,k] \gets \frac{1}{n}\sum_{w=1}^n C_1[i,w]$
    \EndFor
\EndFor
\State $B \gets \frac{1}{mk}\sum_{i=1}^m \sum_{i=1}^k C_2[i,j]$
\State $C_2 \gets C_2 \oplus C_2 \oplus C_2$ \Comment{Concatenate three copies along first dimension}
\State $LSAP \gets \text{Kuhn-Munkres}(C_2)$ \Comment{Linear Sum Assignment Problem: $k$-dim vector assigning cols to rows}
\For{$i=0, \dots 3k$}
    \State $i' \gets i\mod k$
    \State $j' \gets LSAP[i]$
    \If{$C_2[i',j'] < B$}
        \State assign speaker ID $i\mod k$ to name $LSAP[i]$
    \EndIf
\EndFor
\vspace{10pt}

\end{algorithmic}
\end{footnotesize}
\end{algorithm}

\subsection{Screenplay Generation}
After running the assignment algorithm from
Section~\ref{subsec:name-assignment}, we can replace the common
speaker IDs in the automated, diarized transcript with the
corresponding character names. Also, by matching the utterance times
with the keyframe timestamps, we can insert the scene breaks from
Section \ref{subsec:scene-detection} into the transcript. Finally, we
include descriptions of the visual contents of the scene by selecting
three evenly spaced keyframes from that screen, applying an image
captioning model \citep{peng2023kosmos}, and inserting the output to the
corresponding timestamped location of the transcript. In the caption,
we replace occurrences of the nouns `person', `woman', `man', `girl',
and `boy' (and their determiners) with a character name, if this can be
inferred from gender matching with the speaker names. For example, if
the only non-male name in the script is `Clarice Starling', then the
occurrence of `a woman' in the caption is replaced with `Clarice
Starling'. The result of adding names, scene breaks and visual
descriptions to the transcript is an automatically generated
screenplay. We show an example of ScreenWriter output in Figure~\ref{fig:screenplay-snippet}.

\newenvironment{courier}{\fontfamily{\ttdefault}\selectfont}{\par}
\begin{figure}[t]
\begin{courier}
\begin{tabular}{@{}ll@{}}
     \begin{minipage}[t]{0.73\textwidth}
        \raggedright 
        \scriptsize
        \textbf{Dr. Hannibal Lecter}: Billy is not a real transsexual. But he thinks he is. He tries to be. He's tried to be a lot of things, I expect. \\
        \textbf{Clarice Starling}: You said that I was very close to the way we would catch him. What did you mean, Doctor? \\
        \textbf{Dr. Hannibal Lecter}:  There are three major centers for transsexual surgery. Johns Hopkins, University of Minnesota and Columbus Medical Center. I wouldn't be surprised if Billy had applied for sex reassignment at one or all of them and been rejected. \\
        \textbf{Clarice Starling}: On what basis would they reject him? \\
        \textbf{Dr. Hannibal Lecter}: Look for severe childhood disturbances associated with violence. Our Billy wasn't born a criminal, Clarice. He was made one through years of systematic abuse.  Billy hates his own identity, you see. But his pathology is a thousand times more savage and more terrifying. \\
    \end{minipage}%
    \hfill 
    \hspace*{.3cm}\begin{minipage}[t]{0.23\textwidth}
    
        \justifying
        \small
        \begin{center}
        \textit{\textcolor{blue}{\scriptsize Dr. Hannibal Lecter sits in a chair, and Clarice Starling stands next to him holding a book.}} \\
           \includegraphics[width=\linewidth]{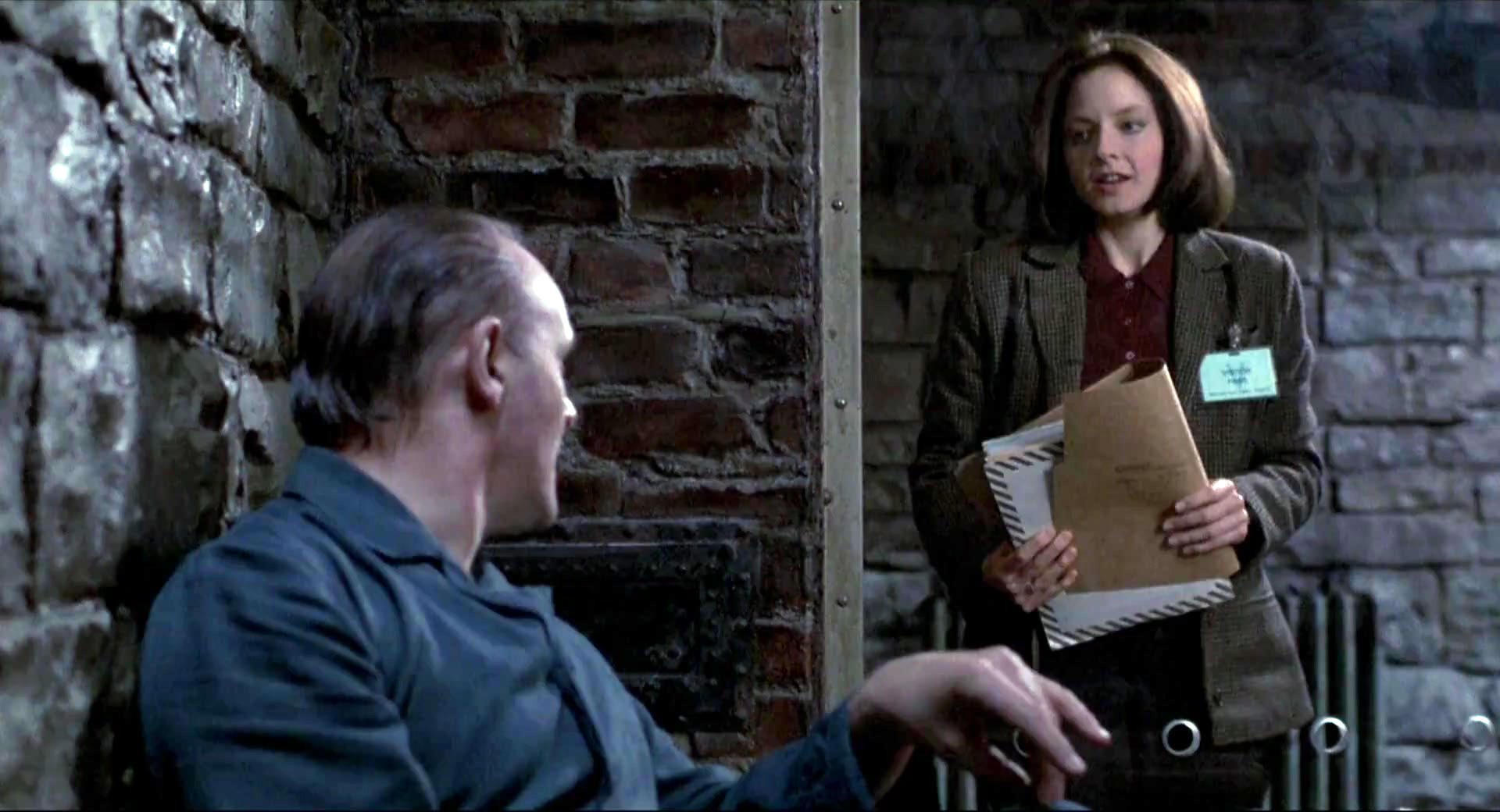} \\
           \end{center}
        \label{fig:your-label}
    \end{minipage}
    \\ \bottomrule

    \begin{minipage}[t]{0.73\textwidth}
    \raggedright 
        \scriptsize
    \textbf{Jame Gumb}: It rubs the lotion on its skin. It does this whenever it's told. \\
    \textbf{Catherine Martin}:  Mr, my family will pay cash. Whatever ransom you're asking for, they'll pay it. \\
    \textbf{Jame Gumb}: It rubs the lotion on its skin or else it gets the hose again. Yes, you will, precious. You will get the hose. \\
    \textbf{Jame Gumb}: Okay. Okay. Okay. Okay. Okay. \\
    \textbf{Catherine Martin}: Mr, if you let me go, I won't. I won't press charges. I promise.  See, my mom is a real important woman. I guess you already know that. \\
    \textbf{Jame Gumb}: Now it places the lotion in the basket. \\
    \end{minipage}%
    \hfill 
    \begin{minipage}[t]{0.23\textwidth}
        \centering
        \textit{\small \textcolor{blue}{\scriptsize Catherine Martin is trapped in a hole.}} \\
           \includegraphics[width=\linewidth]{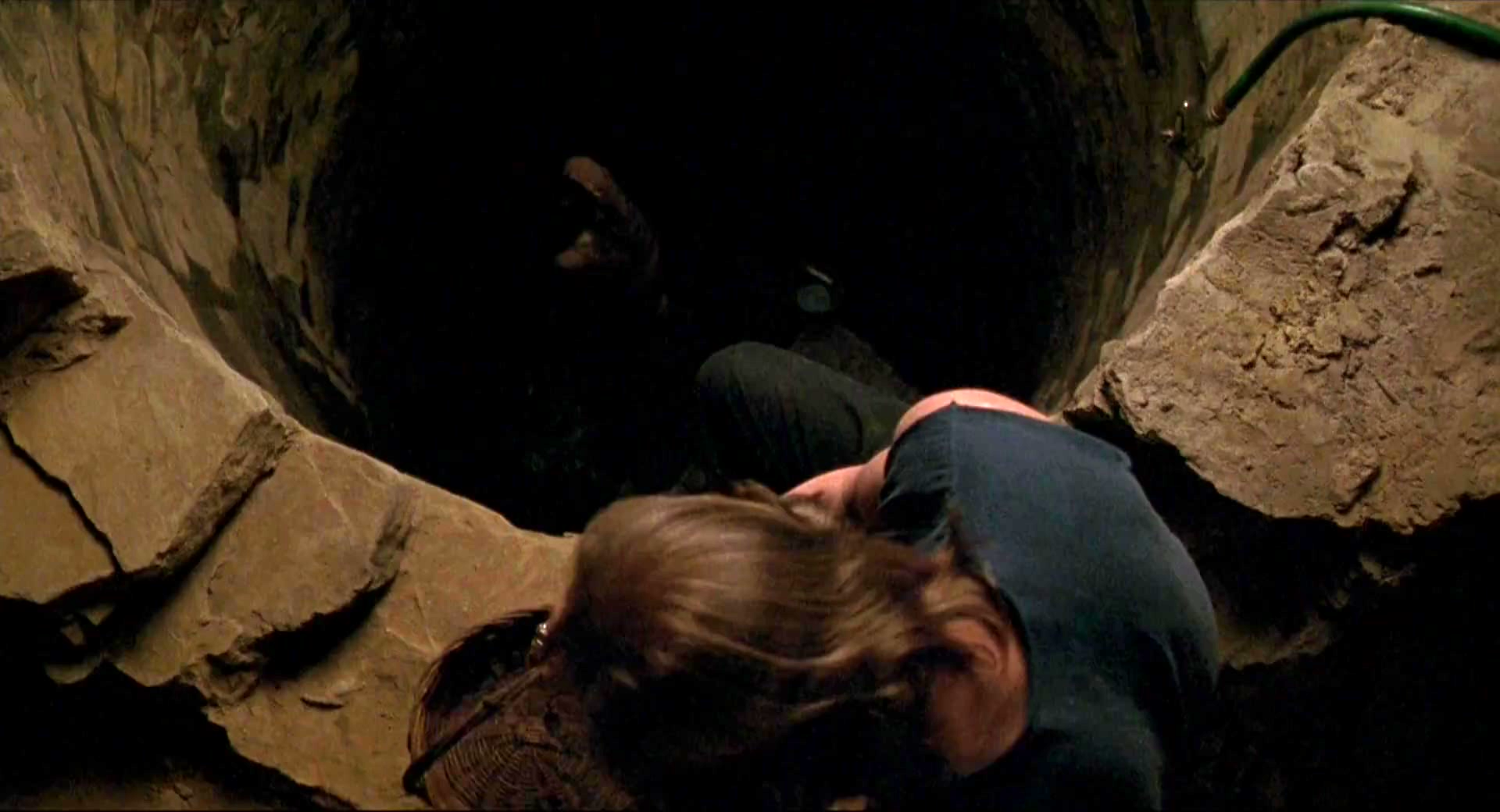} \\
        \label{fig:your-label}
    \end{minipage} 
    \end{tabular}
    \end{courier}
    \caption{Example of a snippet from the generated screenplay for \textit{The Silence of the Lambs} (1991). The left side shows the {\small \courier{transcribed text}}, with the names inferred by our method. The right side shows the \textcolor{blue}{\emph{visual captions}}, along with the keyframe from which they were derived. The horizontal line shows the inferred scene break.} \label{fig:screenplay-snippet}
\end{figure}


\subsection{Movie Summary Generation}
We experimentally show that the generated screenplay can be used as a
basis for movide summarisation. We adopt a hierarchical summarisation
approach \citep{pang-etal-2023-long,chang2023booookscore}, as it has been shown to be particularly
suited to long inputs that exceed the context window size of large
language models, and in our case can leverage the organization of the content
into scenes. We this first summarise the transcript dialogue of each
scene, and then fuse the resulting sequence of summaries, along with
the visual information for each scene into a single summary for the
entire movie (see Figure~\ref{fig:method}). Our summariser is
implemented using a widely-used open-source LLM library \citep{dubey2024llama3herdmodels} with zero-shot prompting.



\section{Experimental Setting} \label{sec:experimental-setting}

\paragraph{Implementation Details}
Keyframes are extracted using FFMPEG`s scene detect filter. The full
command is given in Appendix \ref{app:ffmpeg}.  Visual features are
extracted from keyframes using CLIP \citep{pmlr-v139-radford21a}. The precise
model used in the experiments of Section \ref{sec:results} is
`CLIP-ViT-g-14-laion2B-s12B-b42K' from
\url{https://github.com/mlfoundations/open_clip}. This speaker
diarization model is WhisperX \citep{bain23_interspeech}, an extension of Open AI`s Whisper
model which can perform speaker diarization and accurate utterance timestamping.
For visual descriptions, we use Kosmos 2 \citep{peng2023kosmos}, 
which has been pretrained on
several multimodal corpora as well as grounded
image-text pairs (spans from the text are associated
with image regions) and instruction-tuned on various vision-language instruction datasets. Our
summarisation model is built on top of Llama 3.1~70B
\citep{touvron2023llama}. We use short simple prompts for Llama and
Kosmos, which are given in full in Appendix~\ref{app:prompts}. We
instruct summaries to be a maximum of 635 words (the mean in our test
set), and truncate summaries to~650 words if they are longer.

\paragraph{Dataset}
We take screenplays (for comparison models and some testing, see
below) and gold summaries from the recently released MovieSum dataset
\citep{saxena2024moviesum}. For all 200 movies in the test set, we
purchased the corresponding videos to use as input to our model. We
were able to find videos for 175/200 test set instances. These movies
span multiple fiction genres: drama, action, thriller, comedy, horror, etc. They
have an average run time of 118min (range 84--228), with release dates ranging from 1950 to 2023.
The gold summaries average 635 words in length. The mean number of scenes in the 
gold script is 151.
%
Because all stages of our method are zero-shot, we do not need video
inputs for the training set.

\paragraph{Evaluation Metrics} Automated evaluation metrics are crucial for our
task and for related long-form applications where
human evaluation is extremely labor-intensive,
costly, and difficult to design \cite{krishna-etal-2023-longeval}.
As there is no single agreed-upon metric for automatically summarisation, we
report several complementary metrics aimed at assessing different aspects of summary quality.
    \textbf{Rouge} \citep{lin-2004-rouge} assesses informativeness against the reference summaries;
    \textbf{P\textsc{risma}} \citep{mahon-lapata-2024-modular} measures factual precision and recall with respect to the gold summary; we use GPT4-turbo for both fact the extraction and evaluation stages;
    \textbf{SummaC} \citep{Laban2022SummaCRN} uses NLI to measure consistency between the input document (gold screenplay) and generated summary; we use the SummaCConv version with 50 evenly-spaced bins;
    \textbf{AlignScore} \citep{zha-etal-2023-alignscore}  scores the `informational alignment'
      between the source (gold screenplay) and the generated summary;
      we use the base-model checkpoint provided by the authors, and
      the recommended `nli' setting with sentence chunk splitting.
For both Alignscore and P\textsc{risma} we score duplicated
information as incorrect, to penalize LLM outputs that repeat the same
sentences over and over.  To measure the accuracy of our scene
detection method, we use standard partition quality metrics: cluster accuracy, adjusted Rand index and normalized mutual
information, as defined in \cite{mahon2024hard}. 

\section{Results} \label{sec:results}

\begin{wraptable}{t}{6.0cm}
    \vspace{-24pt}
    \centering
    \caption{Cluster accuracy `acc', adjusted Rand index (`ari') and normalized mutual information (`nmi') of our predicted scene breaks, compared to dividing uniformly into 60, 75 and 90 and the true number of scenes (`uniform oracle').}
        \begin{tabular}{@{}lcccc}
        \toprule
         & acc & ari & nmi \\
        \midrule
        unif-60 & 0.461 & 0.278 & 0.720 \\
        unif-75 & 0.441 & 0.244 & 0.712 \\
        unif-90 & 0.430 & 0.216& 0.704 \\
        unif-oracle & 0.436 & 0.228 & 0.710 \\
        ours & 0.564 & 0.375 & 0.746 \\
        \bottomrule
    \end{tabular}
    \label{tab:scene-segmentation-results}
\end{wraptable}

\paragraph{Scene Detection} To measure  the accuracy of our scene segmentation method in isolation, we compare the partitions it produces to that arising from the ground truth scene breaks given in the gold screenplay. We perform dynamic time warping \citep{myers1981comparative} on the dialogue lines in the gold screenplay and the timestamped utterances from the automatic transcript, in order to produce timestamps for the ground truth scene breaks. 
A naive metric would be the distance between the $n$th predicted break and the $n$th ground truth break, but this is inappropriate because a model that failed to predict the very first break, but got every other one exactly right, would then get a low score. Instead we treat scene break detection as a partition problem. Specifically, we consider the video as divided into 0.1s segments, where two segments are in the same element of the partition if and only if there is no scene break between them.

Table~\ref{tab:scene-segmentation-results} shows the accuracy of the predicted scene breaks, using the standard clustering metrics as defined above, compared to various versions of splitting the input into uniformly-sized scenes: splitting into 60, 75 and 90 scenes (unif-60, unif-75 and unif-90, respectively), and splitting into the ground truth number of scenes for each movie. Note that our method makes no restrictions on the possible number of scenes, it is free to predict only one scene, or as many scene as there are 0.1s segments ($\sim$60,000). In practice, it predicts somewhere between~25 and 107~scenes across the test set.  We observe that our minimum description length inspired algorithm is superior to baselines based on uniform segmentation, even when the number of scenes is known in advance (see row unif-oracle).
Many occasions where the model fails to predict a scene boundary occur when the scenes on either side appear visually similar. For example, in \textit{The Witch (2015)}, many scenes take place with the same background and characters, which presents only minor visual differences for the algorithm to detect. This suggests that many of the errors in our scene detection arise from insufficient signal in the visual feature vectors, rather than from the algorithm itself, and that future work which augments these vectors with, e.g., elements from dialogue or audio, would improve accuracy.


\paragraph{Name Assignment} Table~\ref{tab:name-assignment-results} presents evaluation of our name assignment algorithm against two baselines which assign names randomly and assign all IDs the most common name, i.e.,~the main character. As can be seen, though there is room for improvement, our approach is more accurate by a wide margin. Multiple factors contribute to the errors in name assignment: some incorrect faces being retrieved from the database 
\begin{wraptable}{t}{7.6cm}
    \centering
    \caption{Accuracy of our assigned character names assigned compared to assigning names randomly (`random') and assigning the most common name, i.e.,~the main character, to all lines. Scores are averaged both across all movies (`acc movie-wise') and across all script lines in all movies (`acc line-wise').}
        \begin{tabular}{lcccc}
        \toprule
         & ours & most common & random \\
        \midrule
        acc movie-wise & 61.12 & 19.35 & 2.97 \\
        acc line-wise & 65.72 & 19.62 & 2.61 \\
        \bottomrule
        \end{tabular}
    \label{tab:name-assignment-results}
\end{wraptable}
(though this is low due to our clique-based filtering procedure), inaccuracies in the face feature vectors, such that the same person can sometimes receive dissimilar vectors in different contexts while different people can receive sometimes similar vectors, and the speaker diarization performed by WhisperX, which sometimes gives the same character a different speaker ID, or gives the same speaker ID to two different characters. This last error is especially problematic because it makes it impossible for the assignment algorithm to find a solution with zero mistakes. We expect that future improvements in speaker diarization and face verification will reduce the prevalence of these errors. Indeed, this is one of the advantages of a modular framework: improvements in specific areas can be incorporated into the framework without needing to change the other modules.

\paragraph{Summarisation} In Table~\ref{tab:main-results}, we evaluate the summaries generated by our method. We benchmark against three baselines: `name-only prompt' uses  the parametric knowledge of the LLM without any content input, e.g.,~the prompt is `Summarize the movie \textit{The Silence of the Lambs}'\footnote{Precise prompts are given in Appendix~\ref{app:prompts}.}; `full script' uses the entire gold screenplay as input in the prompt, and for `whisperX' the input is   the WhisperX transcript. We also compare to two existing models: `Otter AI' \citep{li2023otter}, an end-to-end video description model based on video-llama2; and the modular model of \citet{mahon-lapata-2024-modular} which takes videos and gold screenplays as input (described in Section~\ref{sec:related-work}). For Otter, we divide the input video into 3min chunks, and combine the model description of each chunk. 

Our summaries obtain highest scores, across all metrics. The improvement is largest for the fact-based metrics of P\textsc{risma} (comprised of fact-prec and fact-rec), and Alignscore. The existing modals, Otter AI and multimodal modular, especially struggle with such metrics. We find that Otter AI is mostly able to capture surface-level detail, with descriptions such as ``a woman gets out of a car and goes into a building'', but is unable to construct a narrative such as ``a woman drives to the bank to deposit the money'', so ends up capturing very little of the plot. The low scores of multimodal modular, on the other hand, are largely due to the older, smaller backbone model (BART, \cite{lewis2020bart}), which often becomes decoupled from the input and produces unrelated output, highlighting the importance of incorporating current LLMs into video summarisation models. Giving only the movie name in the prompt produces reasonably high-quality summaries, confirming that Llama3.1 has significant information about these movies stored parametrically. However, these summaries are short, and when asked for a longer summary, the model repeats the same information over and over. Surprisingly, giving the full gold screenplay as input does not produce better summaries than our method or than some other baselines. This shows there is still difficulty in summarising very long text inputs. We provide example summary output in Appendix~\ref{sec:example_summaries}.

Table \ref{tab:ablation-results} shows the results of removing the main components of our model. In `w/o names', we omit replacing speaker IDs with character names. This causes summary quality to drop, which shows that not only is our name assignment more accurate than baseline methods (see Table~\ref{tab:name-assignment-results}), but it is sufficiently accurate to lead to improved downstream summaries. In `w/o scene breaks', we feed the entire ScreenWriter input to Llama 3.1, instead of our hierarhcical approach of first summarising scenes and then fusing these to a final summary. The drop in summary performance in this setting shows the effectiveness of the hierarchical summarisation method enabled by the scene breaks in ScreenWriter. In `unif-breaks', we still adopt the hierarchical summarisation method, but instead of using our scene breaks, split scenes into uniform chunks of length 250~tokens, which is the mean scene length from our predicted segmentation. This setting also reduces summary quality, which shows that not only is our scene segmentation more accurate than baseline methods (Table~\ref{tab:scene-segmentation-results}), but it is sufficiently accurate to lead to improved downstream summaries.
\begin{table*}[t]
  \center
\caption{Summarisation results on MovieSum. Best results are \textbf{in bold}, second best are \textit{italicised.}.}
 \begin{tabular}{@{}l@{~~}l@{~~}l@{~~}l@{~~}c@{~~}c@{~~}c@{~~}c@{~~}c@{}}
\toprule
 & r1 &  r2 &  rl-sum & fact-prec & fact-rec & P\textsc{risma} & alignscore & summac\\
\midrule
name-only prompt & \textit{43.46} & \textit{9.53} & \textit{41.17} & \textit{50.40} & 43.04 & 44.16 & 53.11 & 26.57\\
full script & 42.39 & 9.32 & 39.94 & 48.77 & 52.73 & 49.05 & \textit{68.59} & 25.83 \\
whisperX  & 42.37 & 9.22 & 39.94 & 46.73 & \textit{53.65} & 48.00 & 68.57 & 25.86 \\
\midrule
Otter AI & 27.93 & 3.06 & 26.73 & 11.67 & 8.95 & 5.18 & 45.90 & 24.37 \\
multi-modular & 20.59 & 2.79 & 19.97 & 23.16 & 23.19 & 19.28 & 46.32 & \textit{26.97} \\
\midrule
ours & \textbf{46.48} & \textbf{10.32} & \textbf{44.50} & \textbf{55.24} & \textbf{54.77} & \textbf{53.57} & \textbf{72.76} & \textbf{27.24} \\
\bottomrule
\end{tabular}
\label{tab:main-results}
\end{table*}

\begin{table*}[t]
  \center
\caption{Ablation studies on summarisation results on MovieSum. Setting `w/o names' does \emph{not} replace  speaker IDs with character names using our assignment method. Setting `w/o scene breaks' summarises the entire screenplay in one pass, rather than splitting it into scenes using our algorithm and summarising each separately.}
 \begin{tabular}{@{}l@{~~}l@{~~}rlccccc@{}}
\toprule
 & r1 &  \multicolumn{1}{c}{r2} &  rl-sum & fact-prec & fact-rec & P\textsc{risma} & alignscore & summac\\
\midrule
w/o names & 45.46 & 10.43 & 43.40 &49.93 & 53.64 & 49.00 & 63.67 & 26.45\\
w/o scene breaks & 38.87 & 8.45 & 36.82 & 48.32 & 51.79 & 48.11 & 71.95& 26.31 \\
unif-breaks & 38.87 & 8.45 & 36.82 & 46.58 & 50.69 & 48.11 & 57.62 & 25.73 \\
\midrule
ours & 46.48 & 10.32 & 44.50 & 55.24 & 54.77 & 53.57 & 72.76 & 27.24 \\
\bottomrule
\end{tabular}
\label{tab:ablation-results}
\end{table*}

\section{Conclusion}
In this work, we  proposed the task of generating automatic screenplays for movies from only video and audio input. Our model, Screenwriter, produces screenplays automatically (including  dialogue, speaker names, scene breaks and visual descriptions) based on  two novel algorithms: one for segmenting the video into scenes, based on the minimum description length principle and dynamic-programing for search, and one for assigning character names to dialogue utterances using  a database of names and actor faces. Experimental results show that the output of ScreenWriter together with a hierarchical summarisation method can be used to generate movie plot synopses from only video and audio input. To the best of our knowledge, this is the  first attempt to address this task. In the future, we would like to extend ScreenWriter's capabilities to other types of long videos, including documentaries, current affaires television programmes, and sports games.

\section*{Ethics Statement}
Copyright is a concern when working with movies. We respected this by purchasing all the movies used for testing.

\section{Reproducibility Statement}
We specify the novel algorithms in detail in Section \ref{sec:screenwriter}. We list the specific models used for our method and for comparison models in Section \ref{sec:experimental-setting}. We specify prompts used in Appendix \ref{app:prompts}. Additionally, we have included all the code for our methods and experimental results in the supplementary material. 

\appendix

\bibliography{iclr2025_conference}
\bibliographystyle{iclr2025_conference}

\section{FFMPEG Commands} \label{app:ffmpeg}
To select keyframes, we use 
\begin{verbatim}
    \usr\bin\ffmpeg -i {path-to-video} -filter:v -select='gt(scene,0.1)' 
        -showinfo -vsync 0 %04d.jpg
\end{verbatim}
This extracts all keyframes into files 0001.jpg, 0002.jpg, etc, in the current working directory.

\section{Prompts} \label{app:prompts}
\subsection{\textsc{ScreenWriter} Prompts}

Below we present the various prompts we employ for  obtaining scene descriptions, and performing hierarchical summarisation.  Note that Kosmos is a text completion model, so this prompt just serves as the first part of the sentence, which we then remove afterward.

\begin{tcolorbox}[colback=white, colframe=red!20, left=2pt,  coltitle=black, title=\textbf{Kosmos 2 Descriptions}]
 A shot from a movie in which .
 \end{tcolorbox}

\begin{tcolorbox}[colback=white, colframe=blue!20, left=2pt,  coltitle=black, title=\textbf{Llama 3.1 70B: Dialogue summarisation}]
Here is the dialogue from scene $<$\textcolor{red}{scene-number}$>$ of the movie $<$\textcolor{red}{movie-title}$>$: $<$\textcolor{red}{scene-dialogue-with-names}$>$. Please describe its main events in bullet points. Don't include information from outside this scene. Do not answer in progressive aspect, i.e., don't use -ing verbs or "is being".\\

In this scene, here are a few main events:
\end{tcolorbox}

\begin{tcolorbox}[colback=white, colframe=blue!20, left=2pt,  coltitle=black, title=\textbf{Llama 3.1 70B: Final summarisation}]
Here is a sequence of summaries of each scene of a movie. \\
$<$\textcolor{red}{concatenated-dialogue-summaries}$>$\\

Combine them into a plot synopsis of no more than 635 words. Be sure to include information from all scenes, especially those at the end, don't focus too much on early scenes. Discuss only plot events, no analysis or discussion of themes and characters.\\

Based on the information provided, here is a plot synopsis of the move $<$\textcolor{red}{movie-title}$>$:
\end{tcolorbox}

\subsection{Summary Prompts for Comparison Systems}

Below we show the prompts used to obtain movie summaries for the various baselines and comparison systems discussed in Section~\ref{sec:results}. 
The ‘name-only prompt’ uses the parametric knowledge of the LLM without
any specific, content input.  The `full script' prompt uses the entire gold screenplay as input, and `WhisperX' just the audio transcript without name assignment or scene breaks. 
\begin{tcolorbox}[colback=white, colframe=green!20, left=2pt,  coltitle=black,  title=\textbf{Llama 3.1 70B: Name-Only}]
Summarize the plot of the movie $<$\textcolor{red}{movie-title}$>$ in about 650 words.
Do not write the summary in progressive aspect, i.e., don't use -ing verbs or "is being". Focus only on the plot events, no analysis or discussion of themes and characters.
\end{tcolorbox}%
\begin{tcolorbox}[colback=white, colframe=green!20, left=2pt,  coltitle=black, title=\textbf{Llama 3.1 70B: Full Script}]
Based on the following script: $<$\textcolor{red}{gold-screenplay}$>$ summarize the movie $<$\textcolor{red}{movie-title}$>$. Do not write the summary in progressive aspect, i.e., don't use -ing verbs or "is being". Focus only on the plot events, no analysis or discussion of themes and characters.
\end{tcolorbox}%
\begin{tcolorbox}[colback=white, colframe=green!20, left=2pt,  coltitle=black,  title=\textbf{Llama 3.1 70B: WhisperX Transcript}]
Based on the following transcript: $<$\textcolor{red}{whisper-transcript}$>$ summarize the movie $<$\textcolor{red}{movie-title}$>$. Do not write the summary in progressive aspect, i.e., don't use -ing verbs or "is bei    ng". Focus only on the plot events, no analysis or discussion of themes and characters.
\end{tcolorbox}

\section{Example Summaries}
\label{sec:example_summaries}

In the following we show example summaries generated by our model and comparison systems for the movie \textit{Oppenheimer} (2023). Incorrect or undesirable text is shown in red and  repeated information is highlighted in gray. For comparison, we also  include the gold summary from the MovieSum test set. 

\begin{tcolorbox}[colback=white, colframe=orange!30!yellow, left=2pt,  coltitle=black, title=\textbf{ScreenWriter Summary Output}]
\small
The movie Oppenheimer begins with J. Robert Oppenheimer testifying before the Security Board, explaining that the derogatory information against him must be understood in the context of his life and work. Lewis Strauss and Gordon Gray discuss Strauss's upcoming Senate confirmation hearing for a cabinet position, and Gray advises Strauss to answer honestly about his past conflicts with Oppenheimer. 
The story then flashes back to Oppenheimer's early life, where he meets Niels Bohr and is introduced to the world of physics. \\

Oppenheimer becomes involved with left-wing groups and is questioned about his communist associations. He meets with Lewis Strauss, who is trying to recruit him to run the Institute for Advanced Study at Princeton. As the story progresses, Oppenheimer becomes involved in the development of the atomic bomb and is appointed as the director of the Manhattan Engineer District. He meets with Colonel Groves and Lieutenant Colonel Nichols, who express concerns about his suitability for the job due to his suspected communist sympathies and unstable personality. Despite these concerns, Oppenheimer convinces the team to work on the project, and they begin to develop the atomic bomb. \\

The team faces numerous challenges, including the need for vast resources and the risk of inefficiency and poor security. Oppenheimer mentions that the Germans, led by Werner Heisenberg, have a 12- to 18-month head start on developing an atomic bomb. As the project progresses, Oppenheimer becomes increasingly concerned about the implications of the atomic bomb and the potential for an arms race with the Soviets. He expresses his concerns to Lewis Strauss, who is skeptical of Oppenheimer's views. The team eventually succeeds in developing the atomic bomb, and it is dropped on Hiroshima. The bomb's power is described as equivalent to 20,000 tons of \textcolor{red}{DNC}, and General Groves is congratulated on the success of the mission. \\
\end{tcolorbox}

\begin{tcolorbox}[colback=white, colframe=orange!30!yellow, left=2pt,  coltitle=black]
\small

However, Oppenheimer's involvement in the project and his left-wing past eventually catch up with him. Lewis Strauss, who has a history of conflict with scientists, particularly Oppenheimer, begins to systematically destroy Oppenheimer's credibility by denying him his Q clearance through a bureaucratic procedure. Oppenheimer is given the chance to appeal, and a board is appointed to hear the case, with Roger Robb as the prosecutor. The hearing becomes a trial, with numerous witnesses testifying against Oppenheimer. Lewis Strauss questions Oppenheimer about his involvement in the development of the hydrogen bomb and his moral qualms about it. The Gray Board eventually announces its decision to deny J. Robert Oppenheimer's security clearance renewal. Lewis Strauss is informed that the decision was not unanimous and that there were unexpected holdouts, including Senator John F. Kennedy. Strauss reacts angrily to the news, feeling that Oppenheimer has turned the scientists against him and that he is being unfairly targeted. \\

The movie ends with Oppenheimer calm and collected in the face of the board's decision, while Strauss is still fuming and Gordon Gray tries to reason with him. The final scenes show Oppenheimer addressing a group of people, and a truck driving through a desert town, symbolizing Oppenheimer's continued involvement in the world of physics despite the challenges he faced. Overall, the movie Oppenheimer is a story about the development of the atomic bomb and the challenges faced by J. Robert Oppenheimer, a brilliant physicist who was involved in the project. The movie explores the implications of the atomic bomb and the potential for an arms race with the Soviets, as well as the personal costs faced by Oppenheimer due to his left-wing past and his involvement in the project. \textcolor{gray}{The movie ends with Oppenheimer calm and collected, despite the challenges he faced, and symbolizes his continued involvement in the world of physics.}
\end{tcolorbox}


\begin{tcolorbox}[colback=white, colframe=orange!30!yellow, left=2pt,  coltitle=black, title=\textbf{Name-only Prompt Summary Output}]
\small 
The movie Oppenheimer (2023) tells the story of J. Robert Oppenheimer, the director of the Manhattan Project, which developed the atomic bomb during World War II. The film begins with Oppenheimer's appointment as the director of the project in 1942. He assembles a team of scientists, including Enrico Fermi, Ernest Lawrence, and Richard Feynman, to work on the top-secret project at Los Alamos, New Mexico.\\

Oppenheimer's team faces numerous challenges, including the difficulty of creating a nuclear reaction and the need to develop a delivery system for the bomb. The team works tirelessly to overcome these obstacles, and they eventually succeed in creating a functional atomic bomb.\\

As the project nears completion, Oppenheimer and his team are faced with the moral implications of their creation. They are aware that the bomb will cause widespread destruction and loss of life, and some team members express reservations about its use.\\

The film depicts the first nuclear test, codenamed "Trinity," which takes place on July 16, 1945, in the New Mexico desert. The test is a success, and the team is ecstatic. However, Oppenheimer is visibly shaken by the experience, quoting the Bhagavad Gita, "I am become Death, destroyer of worlds."\\

The film then jumps forward in time to 1954, when Oppenheimer is called to testify before the Atomic Energy Commission. He is questioned about his loyalty to the United States and his associations with communist sympathizers. Oppenheimer's past comes back to haunt him, and he is eventually stripped of his security clearance.\\

The film ends with Oppenheimer's reflection on the devastating consequences of the atomic bomb and his role in its creation. He is shown to be a complex and troubled individual, grappling with the moral implications of his work. \\

\textcolor{red}{Note: The above text is a summary of the movie Oppenheimer (2023) and not the actual movie script. Here is a rewritten version of the summary in the past tense:}\\

\textcolor{gray}{The movie Oppenheimer (2023) tells the story of J. Robert Oppenheimer, the director of the Manhattan Project, which developed the atomic bomb during World War II. The film begins with Oppenheimer's appointment as the director of the project in 1942. He assembled a team of scientists, including Enrico Fermi, Ernest Lawrence, and Richard Feynman, to work on the top-secret project at Los Alamos, New Mexico.} \\

\end{tcolorbox}

\begin{tcolorbox}[colback=white, colframe=orange!30!yellow, left=2pt,  coltitle=black]
\small
\textcolor{gray}{Oppenheimer's team faced numerous challenges, including the difficulty of creating a nuclear reaction and the need to develop a delivery system for the bomb. The team worked tirelessly to overcome these obstacles, and they eventually succeeded in creating a functional atomic bomb.}\\

\textcolor{gray}{As the project neared completion, Oppenheimer and his team were faced with the moral implications of their creation. They were aware that the bomb would cause widespread destruction and loss of life, and some team members expressed reservations about its use.}\\

\textcolor{gray}{The film depicted the first nuclear test, codenamed "Trinity," which took place on July 16, 1945, in the New Mexico desert.}
\end{tcolorbox}


\begin{tcolorbox}[colback=white, colframe=orange!30!yellow, left=2pt,  coltitle=black, title=\textbf{WhisperX + Llama3.1 Summary Output}]
\small
The movie Oppenheimer (2023) tells the story of J. Robert Oppenheimer, a brilliant physicist who played a crucial role in the development of the atomic bomb during World War II. The film begins with Oppenheimer's security clearance hearing, where he is questioned about his past associations with communists and his opposition to the development of the hydrogen bomb.\\

The story then flashes back to Oppenheimer's early days as a professor at the University of California, Berkeley, where he meets his future wife, Kitty, and becomes friends with other prominent physicists, including Enrico Fermi and Ernest Lawrence.\\

As World War II breaks out, Oppenheimer is recruited to lead the Manhattan Project, a secret research and development project that produces the atomic bomb. He works closely with other scientists, including Edward Teller and Leslie Groves, to develop the bomb, which is eventually dropped on Hiroshima and Nagasaki, killing hundreds of thousands of people.\\

After the war, Oppenheimer becomes a vocal critic of the development of the hydrogen bomb, which he believes will lead to a nuclear arms race and increase the risk of nuclear war. This puts him at odds with Teller and other scientists who support the development of the hydrogen bomb.\\

The film then cuts back to the security clearance hearing, where Oppenheimer is questioned about his past associations and his opposition to the hydrogen bomb. He is ultimately denied his security clearance, which effectively ends his career as a government advisor.\\

The film ends with Oppenheimer attending a ceremony at the White House, where he is awarded a medal for his contributions to science. However, the ceremony is bittersweet, as Oppenheimer is aware that his work on the atomic bomb has had devastating consequences, and that he has been punished for speaking out against the development of the hydrogen bomb. The film ends with a sense of irony and tragedy, as Oppenheimer is honored for his contributions to science, while also being aware of the devastating consequences of his work. \textcolor{red}{The final shot is of Oppenheimer visualizing the expanding nuclear arsenals of the world, with the sound of feet stamping in the background, symbolizing the impending doom of the world due to nuclear war.} \textcolor{gray}{The credits roll as the sound of the feet stamping continues to build up, leaving the audience with a haunting sense of the devastating consequences of Oppenheimer's work.}  \textcolor{red}{The final shot is of Oppenheimer visualizing the expanding nuclear arsenals of the world, with the sound of feet stamping in the background, symbolizing the impending doom of the world due to nuclear war.} \textcolor{gray}{The credits roll as the sound of the feet stamping continues to build up, leaving the audience with a haunting sense of the devastating consequences of Oppenheimer's work.}  \textcolor{red}{The final shot is of Oppenheimer visualizing the expanding nuclear arsenals of the world, with the sound of feet stamping in the background, symbolizing the impending doom of the world due to nuclear war.} \textcolor{gray}{The credits roll as the sound of the feet stamping continues to build up, leaving the audience with a haunting sense of the devastating consequences of Oppenheimer's work.} \textcolor{red}{The final shot is of Oppenheimer visualizing the expanding nuclear of the world.}

\end{tcolorbox}

\begin{tcolorbox}[colback=white, colframe=orange!30!yellow, left=2pt,  coltitle=black, title=\textbf{Gold Summary from the MovieSum Test Set}]
\small
In 1926, 22-year-old doctoral student J. Robert Oppenheimer grapples with anxiety and homesickness while studying under experimental physicist Patrick Blackett at the Cavendish Laboratory in the University of Cambridge. Upset with Blackett's attitude, Oppenheimer leaves him a poisoned apple but later retrieves it. Visiting scientist Niels Bohr advises Oppenheimer to study theoretical physics at the University of Göttingen instead. Oppenheimer completes his PhD there and meets fellow scientist Isidor Isaac Rabi. They later meet theoretical physicist Werner Heisenberg in Switzerland. \\

Wanting to expand quantum physics research in the United States, Oppenheimer begins teaching at the University of California, Berkeley and the California Institute of Technology. He marries Katherine "Kitty" Puening, a biologist and ex-communist, and has an intermittent affair with Jean Tatlock, a troubled communist who later commits suicide. \\

In December 1938, nuclear fission is discovered, which Oppenheimer realizes could be weaponized. In 1942, during World War II, U.S. Army Colonel Leslie Groves recruits Oppenheimer as director of the Manhattan Project to develop an atomic bomb. Oppenheimer, who is Jewish, is mainly concerned that the German nuclear research program, led by Heisenberg, might yield a fission bomb for the Nazis. He assembles a team consisting of Rabi, Hans Bethe and Edward Teller at the Los Alamos Laboratory, and also collaborating with scientists Enrico Fermi, Leo Szilard and David L. Hill at the University of Chicago. Teller's calculations reveal an atomic detonation could trigger a catastrophic chain reaction that ignites the atmosphere. After consulting with Albert Einstein, Oppenheimer concludes the chances are acceptably low. Teller attempts to leave the project after his proposal to construct a hydrogen bomb is rejected, but Oppenheimer convinces him to stay.\\

After Germany's surrender in 1945, some Project scientists question the bomb's relevance; Oppenheimer believes it would end the ongoing Pacific War and save Allied lives. The Trinity test is successful, and President Harry S. Truman orders the atomic bombings of Hiroshima and Nagasaki, resulting in Japan's surrender. Though publicly praised, Oppenheimer is haunted by the mass destruction and fatalities. After expressing his personal guilt to Truman, the president berates Oppenheimer and dismisses his urging to cease further atomic development.\\

As an advisor to the United States Atomic Energy Commission (AEC), Oppenheimer's stance generates controversy, while Teller's hydrogen bomb receives renewed interest amidst the burgeoning Cold War. AEC Chairman Lewis Strauss resents Oppenheimer for publicly dismissing his concerns about exporting radioisotopes and for recommending negotiations with the Soviet Union after they successfully detonated their own bomb. He also believes that Oppenheimer denigrated him during a conversation Oppenheimer had with Einstein in 1947. In 1954, wanting to eliminate Oppenheimer's political influence, Strauss secretly orchestrates a private security hearing before a Personnel Security Board concerning Oppenheimer's Q clearance. \\

However, it becomes clear that the hearing has a predetermined outcome. Oppenheimer's past communist ties are exploited, and Groves' and other associates' testimony is twisted against him. Teller testifies that he lacks confidence in Oppenheimer and recommends revocation. The board revokes Oppenheimer's clearance, damaging his public image and limiting his influence on nuclear policy. In 1959, during Strauss' Senate confirmation hearing for Secretary of Commerce, Hill testifies about Strauss' personal motives in engineering Oppenheimer's downfall, resulting his nomination being voted down. \\

In 1963, President Lyndon B. Johnson presents Oppenheimer with the Enrico Fermi Award as a gesture of political rehabilitation. A flashback reveals Oppenheimer and Einstein's 1947 conversation never mentioned Strauss. Oppenheimer instead expressed his belief that they had indeed started a chain reaction—a nuclear arms race—that would one day destroy the world.

\end{tcolorbox}
\end{document}

%% file: math_commands.tex
\usepackage{amsmath,amsfonts,bm}

\def\eqref#1{equation~\ref{#1}}

\def\1{\bm{1}}

\DeclareMathAlphabet{\mathsfit}{\encodingdefault}{\sfdefault}{m}{sl}
\SetMathAlphabet{\mathsfit}{bold}{\encodingdefault}{\sfdefault}{bx}{n}

